\documentclass[conference]{IEEEtran}
\IEEEoverridecommandlockouts

\usepackage{hyperref}
\usepackage{graphicx}
\usepackage{algpseudocode}
\usepackage{gensymb}
\usepackage{amsmath}
\usepackage{float}
\usepackage{multirow}
\usepackage{array}
\usepackage{subcaption}

\title{\LARGE \bf
Real-Time 3D Profiling with RGB-D Mapping in Pipelines Using
Stereo Camera Vision and Structured IR Laser Ring
}

\author{ \centering
Amal Gunatilake\href{https://orcid.org/0000-0001-7304-3472}{$^{1}$\includegraphics[scale=.08]{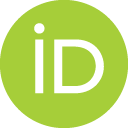}},      
\and
Lasitha Piyathilaka\href{https://orcid.org/0000-0003-4676-9387}{$^{1}$\includegraphics[scale=.08]{orcid.png}},
\and
Sarath Kodagoda\href{https://orcid.org/0000-0001-5175-9138}{$^{1}$\includegraphics[scale=.08]{orcid.png}},
\and
Stephen Barclay$^{2}$,
\and
Dammika Vitanage$^{2}$ 
\and
\parbox{7 in}{\centering 
\fontsize{11}{12}\selectfont $^{1}$iPipes Lab, Centre for Autonomous Systems, University of Technology Sydney, Australia \\
$^{2}$ Sydney Water Corporation, Parramatta, New South Wales, Australia}
}

\begin{document}

\maketitle
\thispagestyle{empty}
\pagestyle{empty}

%%%%%%%%%%%%%%%%%%%%%%%%%%%%%%%%%%%%%%%%%%%%%%%%%%%%%%%%%%%%%%%%%%%%%%%%%%%%%%%%
\begin{abstract}

This paper is focused on delivering a solution that can scan and reconstruct the 3D profile of a pipeline in real-time using a crawler robot. A structured infrared (IR) laser ring projector and a stereo camera system are used to generate the 3D profile of the pipe as the robot moves inside the pipe. The proposed stereo system does not require field calibrations and it is not affected by the lateral movement of the robot, hence capable of producing an accurate 3D map. The wavelength of the IR light source is chosen to be non overlapping with the visible spectrum of the color camera. Hence RGB color values of the depth can be obtained by projecting the 3D map into the color image frame. The proposed system is implemented in Robotic Operating System (ROS) producing real-time RGB-D maps with defects. The defect map exploit differences in ovality enabling real-time identification of structural defects such as surface corrosion in pipe infrastructure. The lab experiments showed the proposed laser profiling system can detect ovality changes of the pipe with millimeter level of accuracy and resolution.

%A unique approach was taken to do 3D laser profiling using stereo IR cameras in combination with structured IR laser. The proposed 3D profiling methodology has many advantages such as getting measurements of the pipe, autonomous identification of defects using algorithms, quality comparison between pipe maintenance and visualization of the data in virtual reality systems. The proposed system is capable of mapping the actual RGB colours seen on the RGB camera when generating the 3D mesh so that it can further contribute to the pipe inspection task by giving the additional colour information of the surface to determine what type of corrosion or damage has happened to the surface. Structured IR laser light with stereo IR cameras were used to improve the accuracy of the laser profiling. Further, it has been used to do the stereo camera triangulation to extract depth information and maneuver the robot inside the pipeline by calculating it’s orientation. Most of the existing systems rely on post-processing the raw data to generate the 3D profile of a pipe. Whereas the proposed system has been implemented to do real-time processing and visualization of the data online by optimizing the algorithms and hardware. Further, it contains data that has been gathered from real sewer pipe sample inspections presented in this paper.

\end{abstract}

%%%%%%%%%%%%%%%%%%%%%%%%%%%%%%%%%%%%%%%%%%%%%%%%%%%%%%%%%%%%%%%%%%%%%%%%%%%%%%%%
\section{INTRODUCTION}

Underground infrastructure such as sewage pipes and water pipes undergo severe concrete \cite{Thiyagarajan2018RobustSewers} and metallic \cite{VallsMiro2018RoboticInspection} corrosion, which considerably reduces their service life. Monitoring such degradation through predictive modeling requires reliable sensor data for high-quality predictions \cite{Thiyagarajan2018RobustInfrastructures, Thiyagarajan2018GaussianInfrastructure, Thiyagarajan2016Data-drivenMeasurementsb,Li2017PredictiveCorrosion}. However, in hostile sewer pipelines, sensors can malfunction over time \cite{Thiyagarajan2018SensorInfrastructures}. In addition to monitoring physical changes of pipes, there are requirements to monitor the sensor health conditions themselves \cite{Thiyagarajan2017PredictiveModel}. Therefore, water utilities around the world are experiencing an uphill battle for maintaining underground assets in a good condition to avoid catastrophic failures such as pipe bursts and ground collapses \cite{Thiyagarajan2016AnTemperature}. Further, human entry to smaller sized pipelines for visual inspections is not possible due to occupational health and safety risks.  Traditionally, CCTV cameras are mounted on remotely operated robotic platforms for inspecting such pipelines, however they only provide visual cues that has limited structural information for decision making.  

In recent years, 3D laser profiling with monocular vision and structured laser light has emerged as a promising technology, which can generate a 3D map of the internal surface of a pipe \cite{Liu2012, Saenz2010, Duran2007, Yoon2009}. The main advantage over CCTV inspection is 3D maps can easily detect and quantify structural defects on the internal pipe surface such as changes in ovality, material loss due to deterioration and or any sludge growths. However, 3D laser profiling using single camera requires extensive field calibrations and the accuracy of the 3D map is largely affected by the lateral movement of the robotic platform. Alternatively, 3D map generation is done mostly offline and therefore, it cannot be used for online as well as opportunistic decision making by using the robotic platform. If a 3D map is available in real-time, then the operation can utilize the 3D data to further investigate the presence of any defects. Further, real-time 3D maps can be used to identify any unscanned regions and improve them accordingly while in operation. 
\begin{figure}[H]
    \centering
    \includegraphics[width=\linewidth]{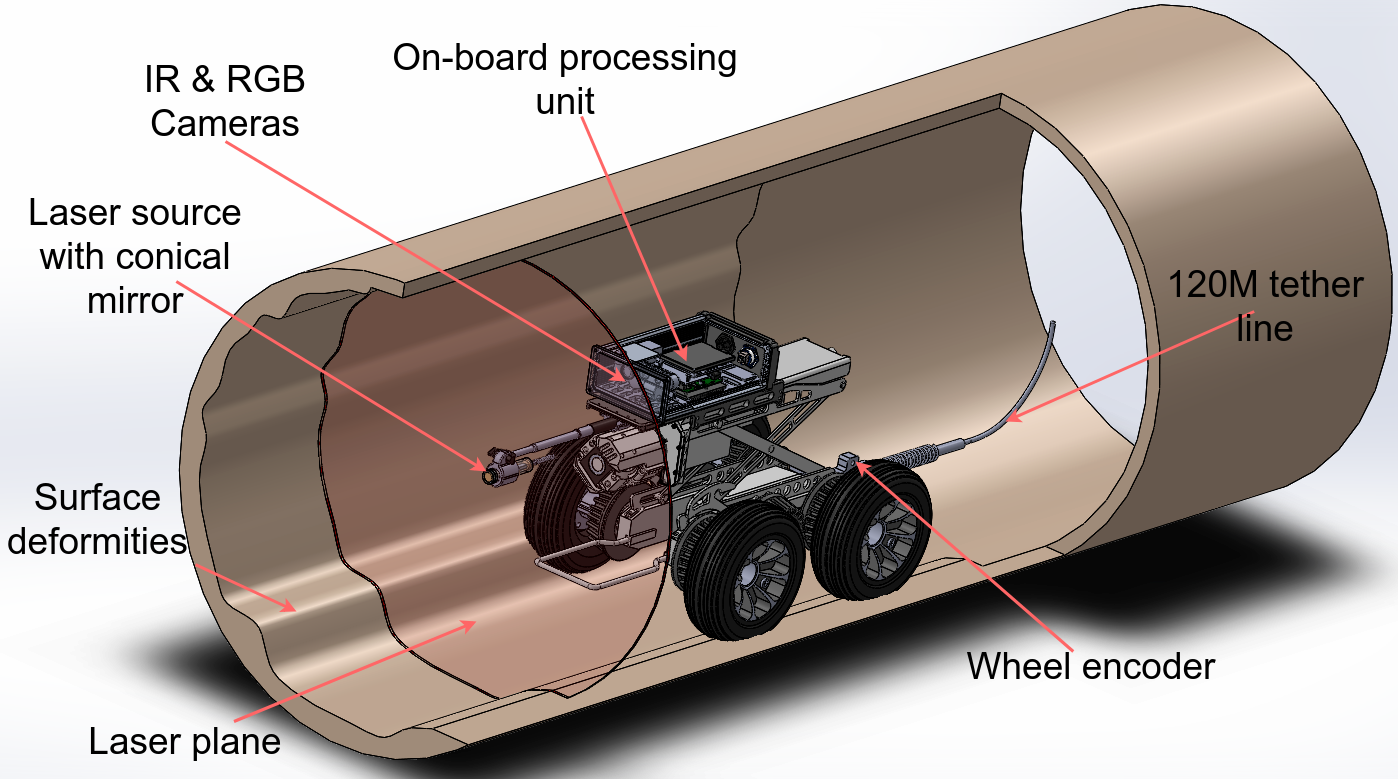}
    \caption{Robot platform with sensors}
    \label{fig:robot}
\end{figure}

This paper proposes a stereo vision-based system that uses IR structured light for 3D profiling inside pipes. The use of stereo cameras enables to decouple the depth measurements of the laser beam from the lateral movement of the robot hence increasing the accuracy of a 3D map. Further, use of stereo vision system does not require field calibration, and can be deployed in a wider range of pipe diameters enabling fast and convenient deployment. The proposed system has the capability to generate the 3D point cloud in real-time while traversing through the pipeline. The real-time 3D map of the pipeline was generated using the ROS framework, and tests were conducted to determine the effectiveness of the 3D maps. An infrared (IR) laser beam projector is proposed in this paper in lieu of traditional red colored laser beam to conserve the true color patterns of the surface on to the 3D map. An additional RGB camera is used to collect the true color data from the surface. The projected IR laser is visible in stereo IR cameras however does not interfere with color values extracted by the RGB camera. In addition, the proposed system generates a color heat map that indicates the deviation of the ovality from the original dimensions of the pipe providing a real-time defect map.

\section{Implementation} 
The sensing prototype was built on a robotic platform, which has the basic functionalities of inspecting pipelines. The platform is equipped with a CCTV camera including a flashlight system that can be deployed via a 120m long tethered cable for connecting with the control station. The stereo camera system and the IR laser projector were built as a separate unit and mounted on the robotic platform with a dedicated computer module as shown in Fig. \ref{fig:robot}.

\subsection{3D profiling technologies}  

Several technologies such as LIDAR, Time-of-flight (TOF) cameras, structured 3D cameras, and structured laser ring profiling were evaluated to identify their performance in generating an accurate 3D map of a pipeline. Among those technologies, LIDAR was less preferred for the proposed application due to low resolution in range measurements (less than 1cm accuracy), and the need of slow operation for increased resolution requirements,
% and for reasonable accuracy, it has to be stationed at a static position while scanning 
which is not ideal for continuous pipe scanning \cite{Nasrollahi2018DesigningInspection}. Although most of the existing 3D cameras are inherent with TOF camera technology, it is less effective due to low grid resolution of the projected pattern. Further, when the surface is reflective the sensing of the projected pattern becomes non trivial and it can result in a considerable  noise \cite{Ujkani2018VisualCell}. Structured light 3D cameras such as  Intel Reaslsense and Microsoft Kinect motion sensor project low resolution IR patterns and therefore they are not sensitive to smaller structural variations of the sensed surface. Since this application demands millimeter (mm) level of accuracy to detect defects on the surface, structured laser ring projection technique with the use of stereo camera vision stands as a strong choice \cite{Stanic2017AProfiler, Lepot2017AQuantification}. This enables to capture accurate structural information from the laser pattern while traversing the robot inside the pipelines \cite{Rantoson2010NonLight, Kofman2007Multiple-lineSensor}. Further research work was conducted in this project to enhance the accuracy and obtain the natural colour patterns of the surface by using a stereo IR camera, a RGB camera and IR laser pattern. Using an IR laser improves the performance of extracting the colour parameters from RGB camera by filtering out the IR light.

\subsection{Hardware} \label{hardware}
A 360 degree IR laser, which  projects  a beam on  to the circumference of the pipe wall was used as the structured light in this setting. The laser projector was mounted at the front-end of the camera as shown in Fig. \ref{fig:robot} to ensure the  laser plane is always projected perpendicular to the robots forward motion as well as to minimize the obstruction.

%This  IR dot laser that  has the same wavelength as IR stereo  cameras (850nm) has been used to project onto a 45\degree angled conical mirror to create a circular shape laser beam. %\The reason behind choosing a ring shape laser beam rather than going for time-of-flight laser projection technology was due to its flexibility and accuracy. The time-of-flight laser grid projection is a hardware calibrated 3D triangulation technique which will make it difficult to make any updates through software in the later stage of testing. Also, since it generates a laser dot grid, when the scan distance increases, the distance between dots become wider and that makes 3D profile resolution low. Whereas laser line projection allows achieving much higher resolution and flexibility to fine tune using software calibration. 
Since the laser projection creates a continuous circular ring throughout the scanning surface area, even small structural changes of the surface are visible as deviations from a perfect circle. Two stereo IR cameras, each  with $1280 \times 720$ resolution, have been used to extract high definition images of the laser projection. Further, an RGB camera with an IR filter, was used to extract the colour information of the pipe wall.
As the chosen IR did not interfere with the RGB spectrum, the 3D point cloud of the IR laser ring generated by the stereo camera can be projected on to the color image to obtain high quality RGB data of the pipe surface. In order to track the robot movements, wheel odometry has been used by fixing a rotary encoder with 1000 pulses/revolution on the robot. Further, to detect the orientation of the robot, stereo camera images were used. In order to run the whole system including image processing algorithms, a high-performance mother board has been incorporated. An Intel motherboard with 2.5GHz Quad core processor, 8GB RAM and 500GB SSD internal storage memory made the system run smooth in real-time. The whole system was built inside a water tight enclosure box to protect the electronics from  water leaks.

\subsection{3D point cloud processing pipeline}
\begin{figure}
    \centering
    \includegraphics[width=\linewidth]{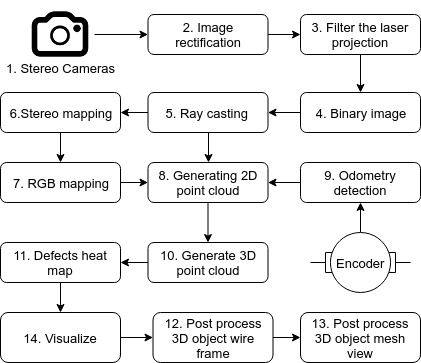}
    \caption{Algorithm operation pipeline}
    \label{fig:block_pipeline}
\end{figure}

A set of standard image processing techniques were utilized to generate a  dense high resolution point cloud in real-time. The execution pipeline of the algorithm is shown in Fig. \ref{fig:block_pipeline}.

% can you divide this section into small sections as follows according to the pipe line - Done
\subsubsection{Camera calibration}

Image distortion is a common problem for any camera. Therefore, in order to get accurate measurements from the camera images, they generally need to be calibrated first using the equation (\ref{eq:distortion1}) and (\ref{eq:distortion2}), where  \(x\) and \(y\) are the  coordinates of the image pixel, \(r\) is the radius from the image center to the pixel, and \(k_1,k_2,k_3\) are the radial distortion coefficients that need to be determined through calibration. 
\begin{eqnarray}
    y_{correction} = y(1 + k_1r^2 + k_2r^4 + k_3r^6 ... + k_nr^{2n}) \label{eq:distortion1} \\*
    x_{correction} = x(1 + k_1r^2 + k_2r^4 + k_3r^6 ... + k_nr^{2n}) \label{eq:distortion2}
\end{eqnarray}
The image distortion factor changes when the pixel coordinates are far from the image center. Therefore, multiple levels of coefficients (\(k_1,k_2,k_3\)) have to be calculated using training data. Most of the cameras satisfy the calibration by estimating the first three coefficients.

When a 3D object is projected onto the camera 2D image plane, the relationship between the 3D coordinates and the 2D plane coordinates can be represented by the camera intrinsic and extrinsic parameters. Equation (\ref{eq:coordinates1}) defines the high level representation of the camera parameters. When \(P_w\) is the 3D coordinates of the object and \(P'\) is the 2D coordinates of the object projected on the image plane, the camera calibration parameters are given by  \(K\), \(R\) and \(T\), where \(K\) is intrinsic parameters and \(R\) (rotation) and \(T\) (translation) are extrinsic parameters. 
\begin{equation}
    P' = [K][R\ T]P_w \label{eq:coordinates1}
\end{equation}
Intrinsic parameters define the image focal length (\(f_x, f_y\)), optical center (\(c_x, c_y\)) and skew coefficients (\(s\)) in pixels. Extrinsic parameters define the relationship of the camera translation (\(T\)) and rotation (\(R\)) within the world coordinates. In practical application of equation (\ref{eq:coordinates1}), it further breaks down into equation (\ref{eq:coordinates2}) where \(X, Y, Z\) are the 3D world coordinates of the object and \(x, y\) represent the coordinates of the object projected on the camera image plane.

\begin{equation}
    \begin{bmatrix}
    x \\ y \\ 1
    \end{bmatrix}
    = 
    \begin{bmatrix}
    f_x \ s \ c_x \\ 
    0 \ f_y \ c_y \\
    0 \ \ 0 \ \ 1 \\ 
    \end{bmatrix}
    \begin{bmatrix}
    r_{11} \ r_{12} \ r_{13} \ t_{1} \\ 
    r_{21} \ r_{22} \ r_{23} \ t_{2} \\
    r_{31} \ r_{32} \ r_{33} \ t_{3} \\
    \end{bmatrix}
    \begin{bmatrix}
    X \\ Y \\ Z \\ 1
    \end{bmatrix} \label{eq:coordinates2}
\end{equation}

Once the algorithm runs through a set of training data set (usually a checker board pattern with known dimensions) the coefficients can be identified. Those parameters can be used for sensor calibration to improve the accuracy of extracting pipe measurements.

\subsubsection{Image masking and filtering} 

The binary mask of the laser circle is extracted from the calibrated image through thresholding. Occasionally, due to surface illumination variations the detected circle had varying line thicknesses, which adversely affected the center detection of the circle. Therefore, to refine this line, a unique filter which had  subtract and bitwise-or image processing functions are implemented using  OpenCV framework. The best fit parameters are determined by training the algorithm using different types of data sets obtained from different pipes under different lighting conditions. 

\subsubsection{Center detection and ray casting} 
The Hough transform algorithm is used to estimate the center of the pipe circle that is used for ray casting and generating a point cloud. The accuracy of the center detection has been much improved by the filter algorithm. By running the ray casting function, all the points related to the laser beam circle can be identified and saved into an array of vector points indexed by the angular rotation. The ray casting algorithm is used mainly to achieve 3 main goals. They are: to generate the point cloud by extracting laser points; stereo mapping to extract colour depth information and to generate a heat map for defect identification by calculating the distance between center to each point. 

\subsubsection{Stereo mapping}
Stereo mapping algorithms are  used to calculate the depth of each point generated from the ray casting algorithm. By using the generated stereo calibration parameters, ray casted points are projected on to the other stereo camera image plane as epipolar lines. In rectified image plane, epipolar lines become parallel horizontal lines. Therefore stereo correspondence process becomes faster and efficient as the search can be done along the raws of the image. The corresponding point can be searched in  the second image by detecting the locations where the laser beam cuts each epipolar line, which can be easily done by searching for highest intensity value. Usually  two  intersection points are detected due to the circular shape of the laser beam. Hence, in order to identify the correct match, the search algorithm can be updated to classify the points according to their location with respect to the center. Equations (\ref{eq:disparity}), (\ref{eq:stereo}) and (\ref{eq:stereo2}) represent the high level calculations of the stereo mapping functionality.

When the disparity value taken as \(d\), camera focal length defined as \(f\), 
distance between two camera focal points defined as \(T\), 
coordinates of the object position defined as \(P = (X_p,Y_p,Z_p)\),
object coordinates in the left camera image defined as \(p_l = (x_l,y_l)\) and 
object coordinates in the right camera image defined as \(p_r = (x_r,y_r)\);
the relation between the coordinates can be defined as below;

% this equation need to be left aligned
\begin{eqnarray}
    d = (x_l - x_r) \label{eq:disparity} \hspace{49 mm} \\
    x_l = \frac{X_p f}{Zp} ;\  x_r = \frac{(X_p - T) f}{Zp} ;\ y_l = y_r = \frac{Y_p f}{Z_p} \label{eq:stereo};
\end{eqnarray}

\begin{figure}
    \centering
    \includegraphics[width=\linewidth]{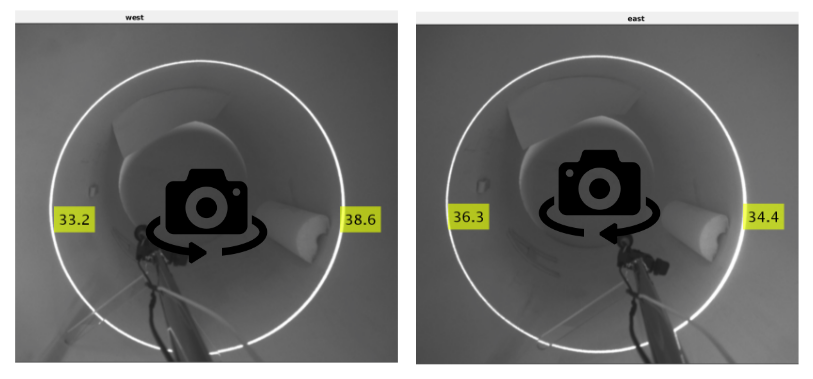}
    \caption{Feature depth calculation for camera tilt detection}
    \label{fig:tilt}
\end{figure}

From (\ref{eq:disparity}) and (\ref{eq:stereo}):
\begin{eqnarray}
    Z_p = f \frac{T}{x_l - x_r} = f \frac{T}{d};\  X_p = x_l \frac{T}{d};\  Y_p = y_l \frac{T}{d} \label{eq:stereo2}
\end{eqnarray}
Once the depth information is available from the stereo mapping algorithm, it can be used to determine the orientation of the robot. If the robot tilt while moving forward the orientations are displayed on the system to maneuver it properly. An example depth calculation on two identified points are shown in Fig. \ref{fig:tilt}. The depth information further can be used to show the pipe orientation in world coordinates. which is planned as a future enhancement.

\subsubsection{RGB mapping}
The generated 3D points are used for  RGB color mapping by projecting them on to the RGB image using the equation (\ref{eq:projection1}) and (\ref{eq:projection2}). When a 3D point (\(P_x, P_y, P_z\)) projects onto 2D point (\(P'_x, P'_y\)) where \(s\) is arbitrary scale factor and \(c\) is arbitrary offset: 
\begin{eqnarray}
    P'_x = s_xP_x + c_x \label{eq:projection1} \\
    P'_y = s_zP_z + c_z \label{eq:projection2}
\end{eqnarray}

Using matrix multiplication, this can be further broken into equation (\ref{eq:projection3}).
\begin{equation}
    \begin{bmatrix}
        P'_x \\ P'_y 
    \end{bmatrix}
    =
    \begin{bmatrix}
        s_x \ 0 \ 0 \\
        \ 0 \ \ 0 \ s_z \\
    \end{bmatrix}
    \begin{bmatrix}
        P_x \\ P_y \\ P_z
    \end{bmatrix}
    +
    \begin{bmatrix}
        c_x \\ c_z \\
    \end{bmatrix} \label{eq:projection3}
\end{equation}

By projecting  the 3D point cloud on to the RGB camera image, the colour information of each point can be obtained. The projection parameters are estimated by stereo calibrating one of the IR cameras with the RGB camera.The color information is later fused with the point cloud data to generate RGB-D map of the pipe.

\begin{figure}[]
    \centering
    \includegraphics[width=3.3in]{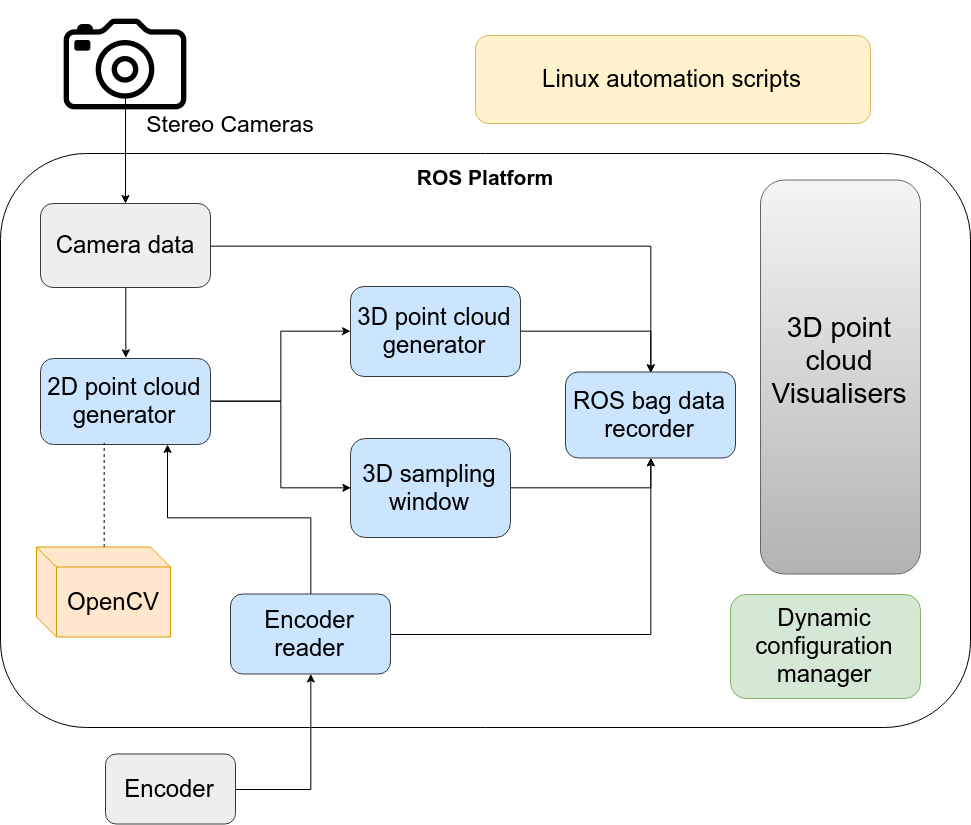}
    \caption{Software architecture diagram}
    \label{fig:component}
\end{figure}

\subsubsection{Software architecture}
The full system was implemented on  Robotic Operating System (ROS) framework using C++ and OpenCV libraries. In order to improve the real-time performances, the major functionalities are distributed into  separate threads by creating appropriate ROS nodes. This approach  enabled the full potential of parallel processing to enhance the performance. The software architecture  is shown in Fig. \ref{fig:component}. The images taken from the cameras are processed in the depth calculation node to generate the 3D point cloud of the laser ring w.r.t camera frame. As the robot moves forward, the odometry information is fused with the 3D point-cloud data of the laser rings to iteratively generate the 3D map of the pipe. By subscribing to each topic using ROS RVIZ like visualization system, the 3D profile map of the pipe can be seen in real-time. Linux automation scripts were employed to automate and start-up process with a single push button.

\section{EXPERIMENTS}
%The tests were conducted on pipe samples in iPipes lab at University of Technology Sydney.

In order to experimentally validate the proposed 3D profiling system, a series of experiments were carried out on various pipe samples.  

\subsection{Accuracy of the diameter measurements }

\begin{figure*}
    \centering
    \includegraphics[height=1.6in]{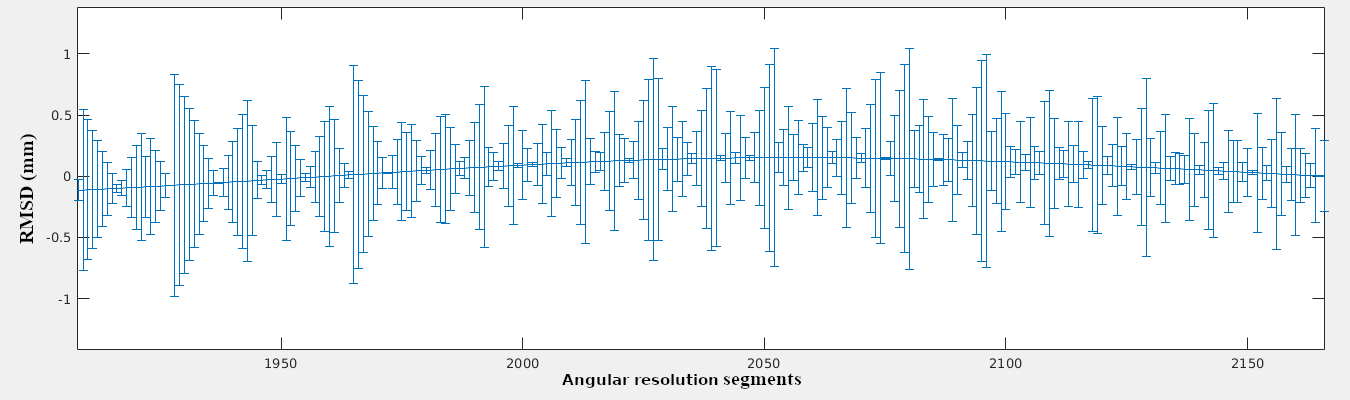}
    \caption{Root mean square deviation from the ground truth}
    \label{fig:error_bar}
\end{figure*}

The ability of the proposed system to measure any ovality change with millimeter level of accuracy is important in applications where accurate defect quantification are needed.  Therefore, the robotic laser profiling was done on a PVC pipe with a diameter of 300mm. The Root Mean Square Error (RMSE) was calculated across the circumference of the pipe for each sample period, and the  test results are shown in Fig.\ref{fig:error_bar}. The  recorded RMSE values were always less than  1mm proving the  ability of the proposed system of measuring 300mm pipe diameter with sub millimeter level of accuracy.

%In order to validate the sensors, a 300mm diameter PVC pipe was used initially. 

%\subsubsection{Diameter validation test}
%First several scans were done to estimate the diameter of the pipe using the system. Then using the standard equations, the root mean square error/displacement (RMSD) metric was calculated by comparing the estimated value with the benchmark value. 
% Equation \ref{eq:rmse} has been used to calculate the error rate. When the root mean square deviation is denoted by \(d'\), the estimated diameter denoted by \(\hat{d}\) and the expected diameter denoted by \(d\). RMSE given by \ref{eq:rmse}. 

% \begin{equation}
%     RMSE (d') = \sqrt{(\hat{d} - d)^2} \label{eq:rmse}
% \end{equation}

\subsection{Quantification of defects}
\begin{figure}
    \centering
    \includegraphics[width=2in]{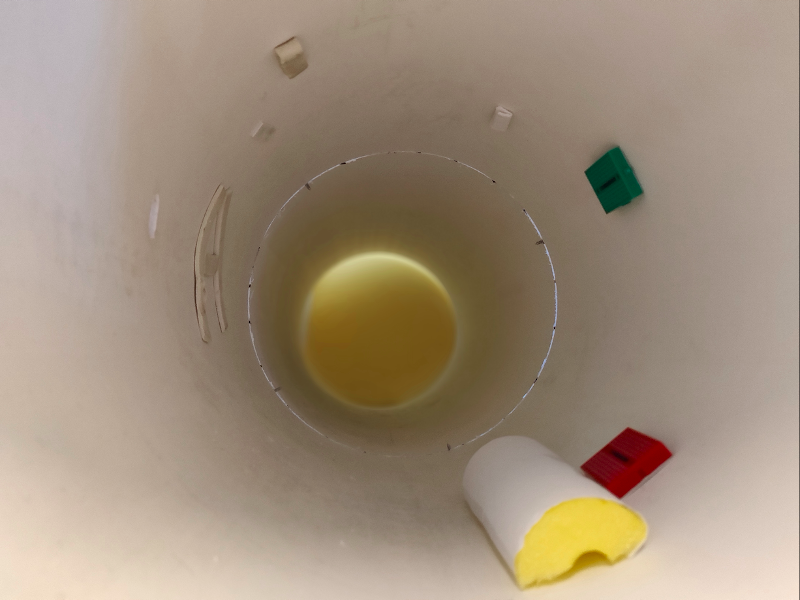}
    \caption{Artificial defects placed on the pipe for accuracy tests}
    \label{fig:defects}
\end{figure}

Next the pipe was modified with artificially planted defects for detection as shown in Fig. \ref{fig:defects}. The thickness of those artificially planted protruded defects were 3mm or more  and some were  artificially made holes of 3mm depth inside the pipe. After running the robotic laser profiling system across the pipe several times,  RMSE values were calculated using  the standard equation. The results are tabulated in Table \ref{tab:testresults}. These test results show the ability of the proposed laser profiling system for  quantifying both protruded and caved in defects with mm level of accuracy. 

%The results from each test were plotted against the angle rotation to check on the deviation from estimated values. Fig. \ref{fig:error_bar} shows a portion of the graph and as it seems, for the 300mm pipe the error rate has been always less than a millimeter. Therefore, it has shown a higher level of accuracy. Based on the tests conducted with different pipe sizes; it has shown the maximum error rate for 600mm diameter pipe is around 2mm.

\begin{table}[H]
    \centering
    \caption{Test results}
    \begin{tabular}{|>{\centering} p{19mm}|>{\centering} m {10mm}|c|c|c|}
        \hline
        \multirow{2}{*}{\textbf{Test}} & \textbf{Ground} & \multicolumn{2}{c|}{\textbf{Result}} & \multirow{2}{*}{\textbf{RMSE}} \\
        \cline{3-4}
         & \textbf{truth} & \textbf{Minimum} & \textbf{Maximum} &  \\
        \hline
        PVC pipe diameter & 300mm & 299mm & 301mm & $\pm$ 1mm \\
        \hline
        Artificial defect 1 (300mm pipe) & 3mm & 2mm & 4mm & $\pm$ 1mm \\
        \hline
        Artificial defect 2 (300mm pipe) & 10mm & 9mm & 11mm & $\pm$ 1mm \\
        \hline
        Artificial holes (300mm pipe) & 3mm & 2mm & 4mm & $\pm$ 1mm \\
        \hline
    \end{tabular}
    \label{tab:testresults}
\end{table}

% Need to divide this section into subsection - Done
%  Sensor Validation- Test on a PVC pipe with known diameter - Done
%  Defect detection- Test on a pipe with artificially created defects. - Done
% test on a corroded cast iron pipe - Done
% real-time performance - Done
% Use  root mean square error. using just mean error is not correct - Done

% Based on the tests conducted with different pipe sizes; it has shown the error rate has a proportionate relationship to the diameter. As an example for the 600mm pipe the error rate increased up to 2mm.

% $\\ \text{300mm pipe} \propto \text{resolution (1280x720)} \propto
% \text{1mm accuracy} \\
% \text{Therefore, since the pixel resolution is fixed,} \\
% \text{600mm pipe} \propto \text{2mm accuracy.} \\$

\begin{figure*}
    \centering
    \begin{subfigure}[b]{0.3\textwidth}
        \includegraphics[width=1.6in]{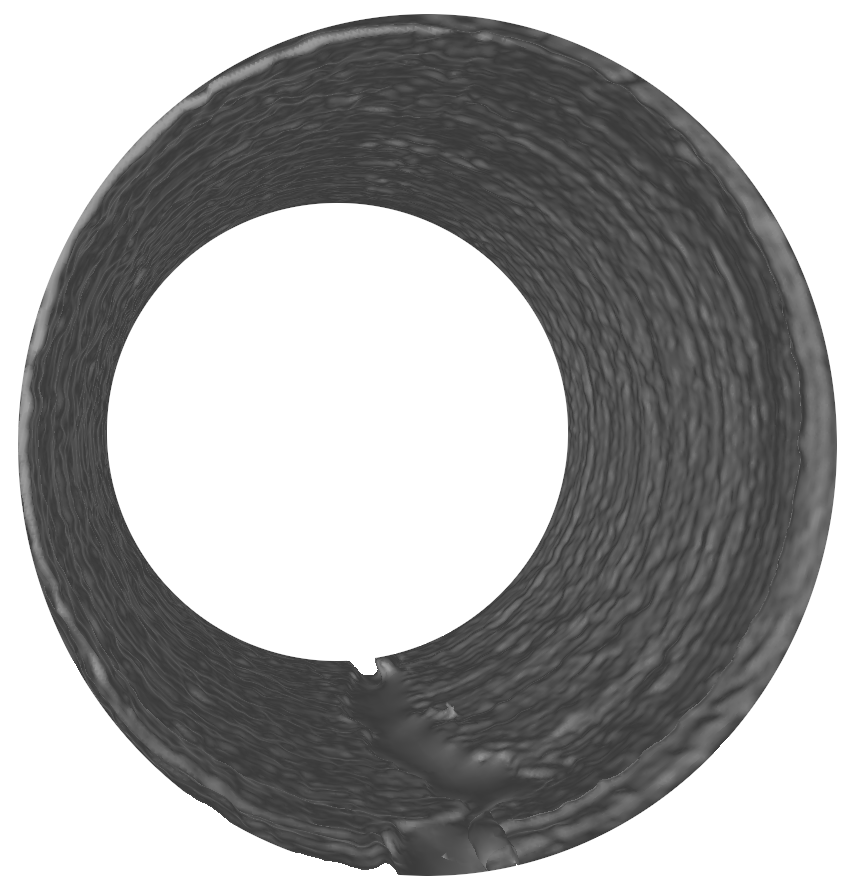}
        \caption{3D mesh of the pipe without colour}
        \label{fig:metal_mesh}
    \end{subfigure}
    \begin{subfigure}[b]{0.3\textwidth}
        \includegraphics[width=1.6in]{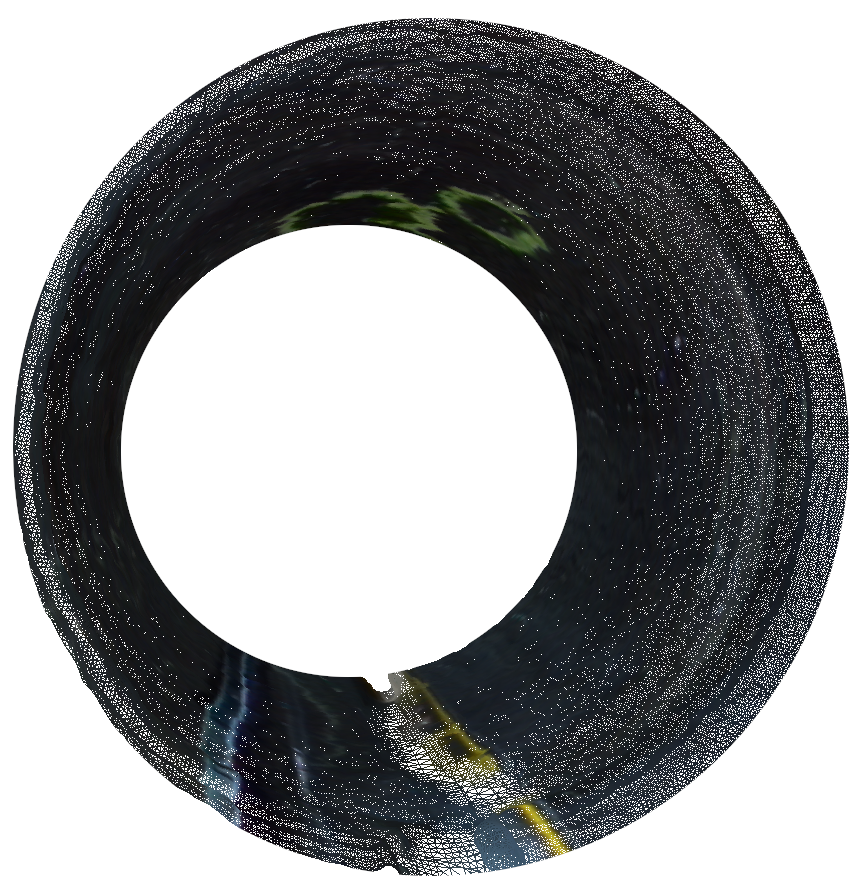}
        \caption{3D wire frame of the pipe}
        \label{fig:rgb_metal_wire}
    \end{subfigure}
    \begin{subfigure}[b]{0.3\textwidth}
        \includegraphics[width=1.6in]{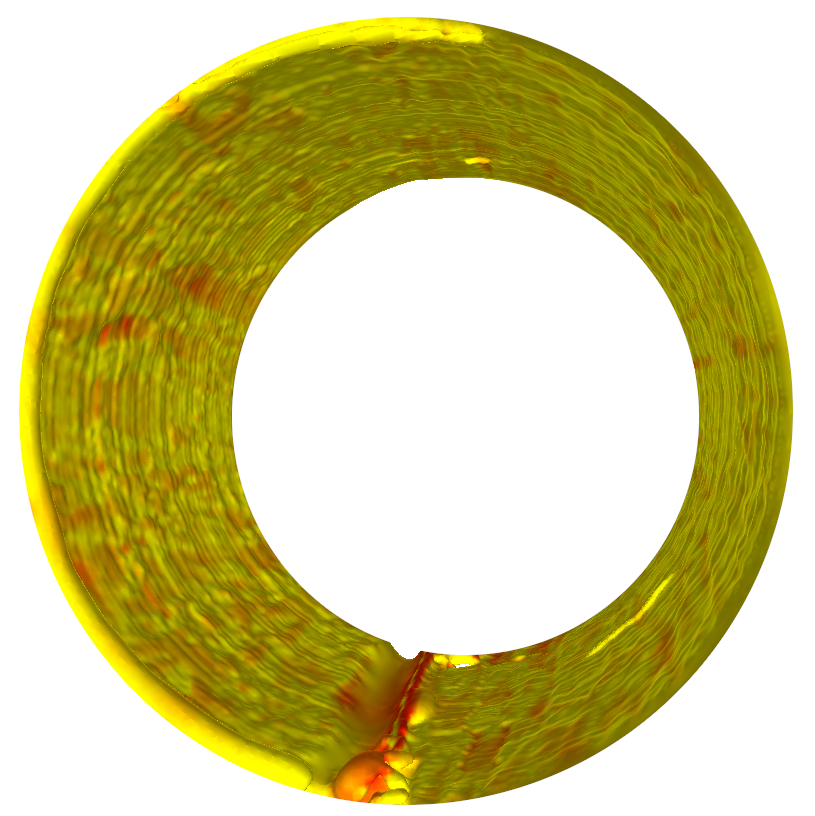}
        \caption{Pipe 3D profile with defects heat map}
        \label{fig:colour_metal_mesh}
    \end{subfigure}
    \begin{subfigure}[b]{0.3\textwidth}
        \includegraphics[width=1.8in]{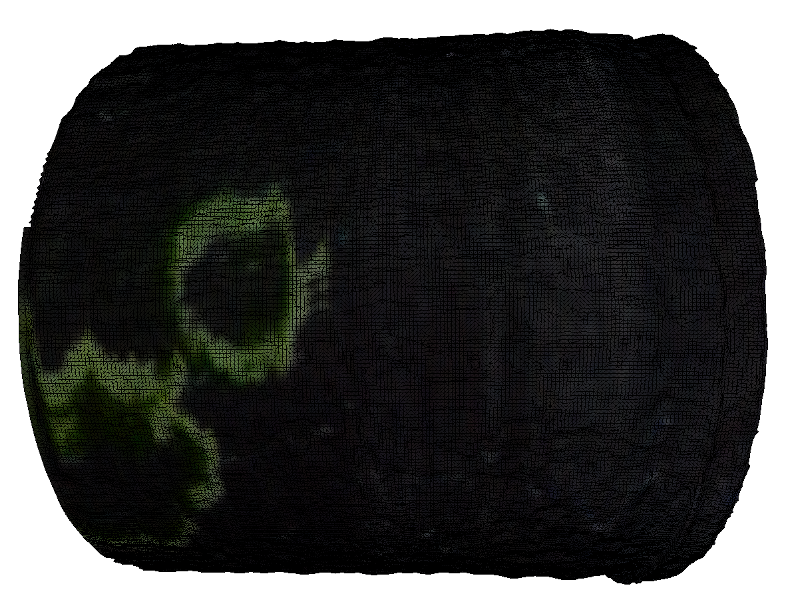}
        \caption{3D mesh view with embedded real colour}
        \label{fig:rgb_metal_mesh}
    \end{subfigure}
    \begin{subfigure}[b]{0.3\textwidth}
        \includegraphics[width=1.6in]{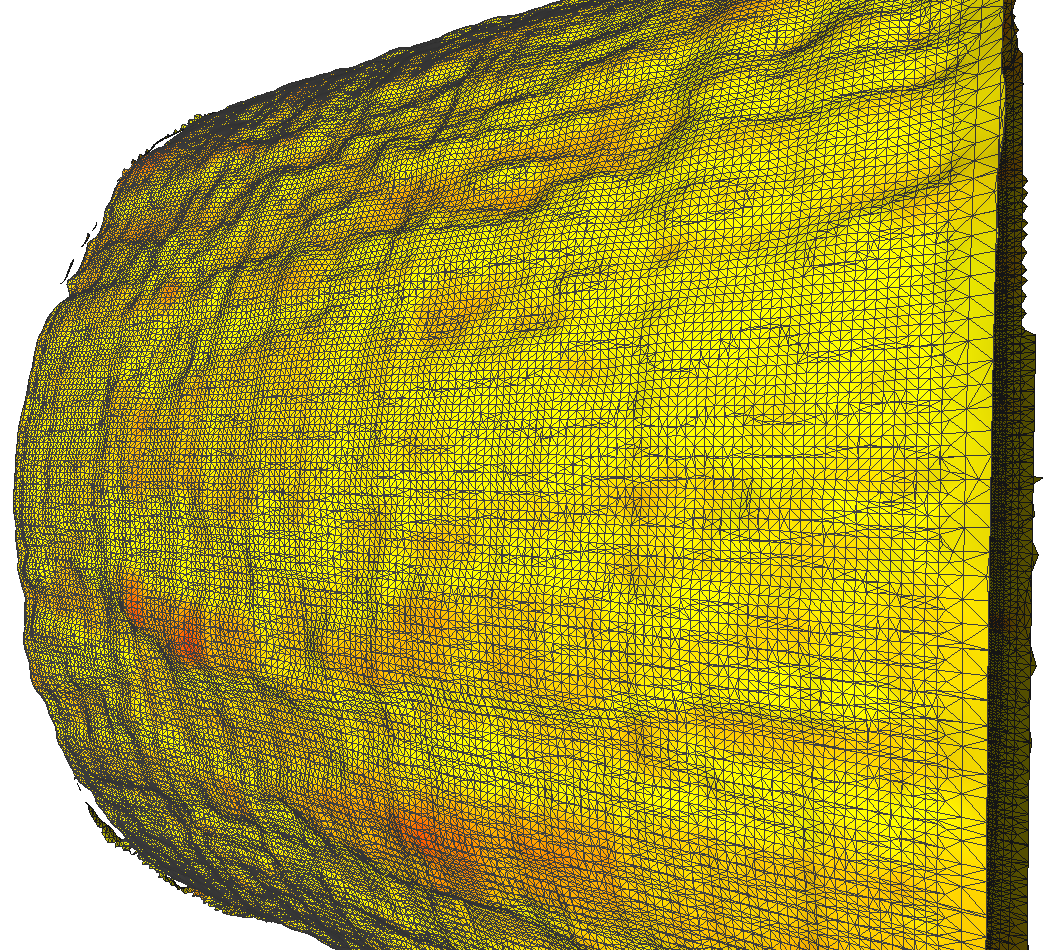}
        \caption{Side view of the defects heat map}
        \label{fig:side-view}
    \end{subfigure}
    \begin{subfigure}[b]{0.3\textwidth}
        \includegraphics[width=1.8in]{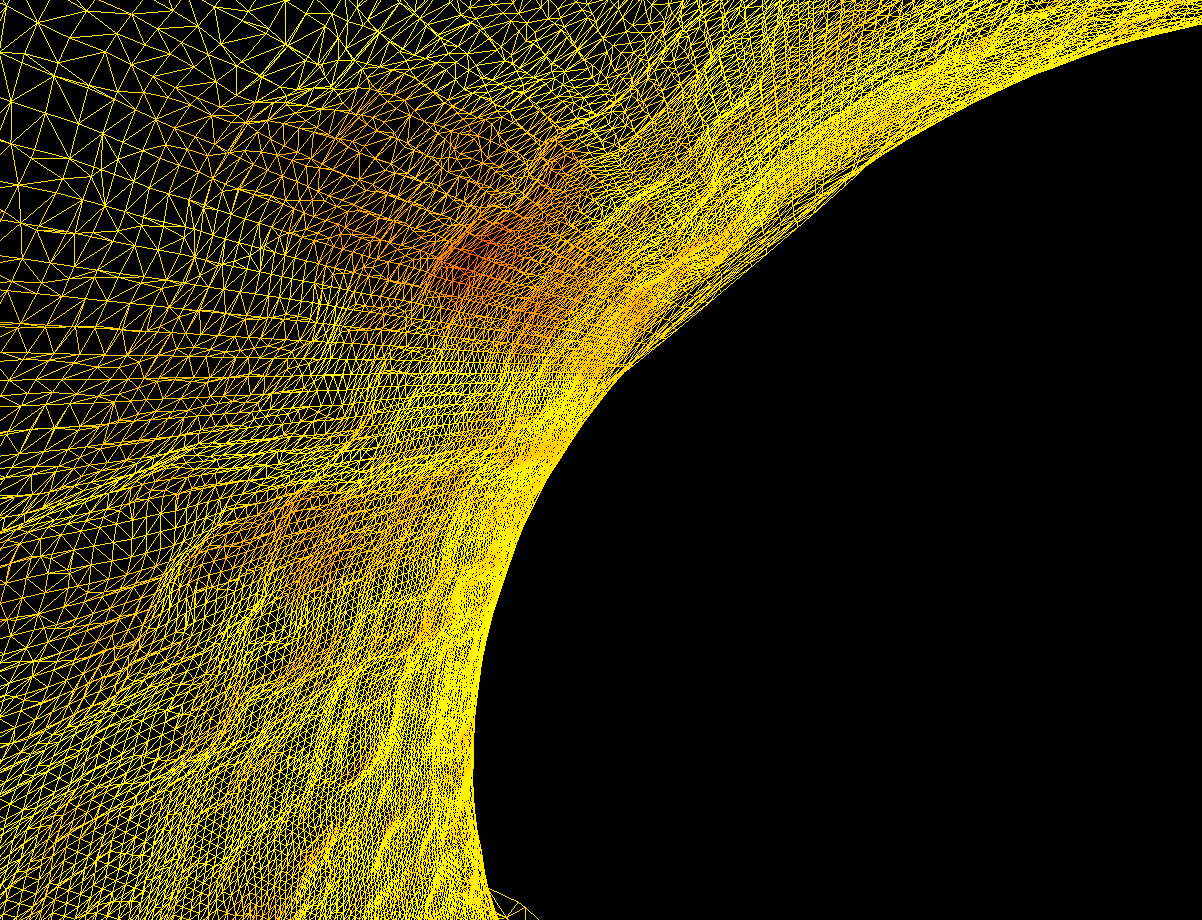}
        \caption{Detailed mesh wire of the heat map}
        \label{fig:colour_metal_zoom_wire}
    \end{subfigure}
    \caption{3D scan results}
    \label{fig:my_label}
\end{figure*}

%\subsubsection{Localization test}
%For the localization odometry values were taken from a rotary encoder which has \(1000 pulses/ revolution\). The circumference of the wheel attached to the rotary encoder is \(200mm\). When robot moves at it's lowest speed the accuracy gained from the odometry localization is around 1mm. 
% The scan resolution used for the ray casting algorithm is 3000 points per revolution. This gives an angular resolution of 0.12\degree. For the odometry a rotary encoder with \(1000 pulses/ revolution\) has been used and the circumference of the wheel attached to the rotary encoder is \(200mm\). Therefore, the odometry resolution is \(0.2mm\). In order to achieve a high-resolution accuracy, the robot was moved in the slowest speed level (\(30mm/s\)) when scanning the pipes. For each second camera takes \(30 frames/second\). Therefore, between each frame the accuracy of the point cloud is \(1mm\) along the Z axis. 

\begin{figure}[H]
    \centering
    \includegraphics[width=2.5in]{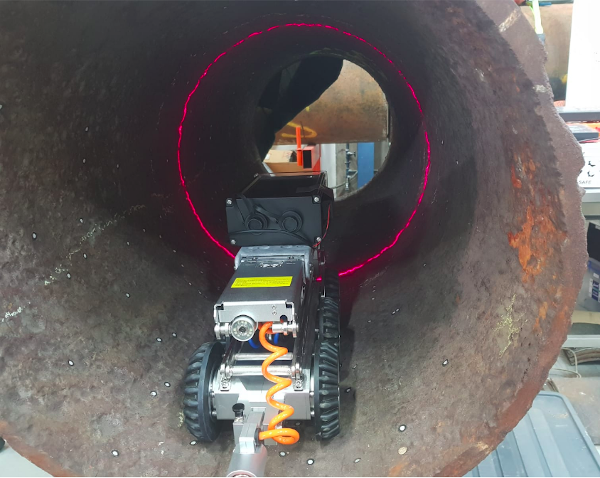}
    \caption{The robot scanning a corroded metal pipe. Red laser projection has been used instead of the IR laser for image demonstration purposes.}
    \label{fig:metal_pipe}
\end{figure}

\subsection{Scanning corroded pipe samples}
After validating the sensors and testing the accuracy, several corroded pipe samples were used to test the entire system. Among them are corroded cast iron  pipes with visible corrosion pits. Fig. \ref{fig:metal_pipe} shows an example of performing a laser profiling in a corroded 600mm metal pipe. The scan results are presented from  Fig. \ref{fig:metal_mesh} to Fig. \ref{fig:colour_metal_zoom_wire} and the defects are highlighted and presented from Fig. \ref{fig:colour_metal_mesh} to Fig. \ref{fig:colour_metal_zoom_wire}. Shades of red colour has been applied to the surface defects according to the corroded depth. The 3D profiling was able to map  corrosion pits  with an accuracy level of 2mm for 600mm pipe, which was verified through spot measurements. Fig. \ref{fig:colour_metal_zoom_wire} shows a magnified portion of the wire frame with highlighted corrosion pits and their depth heat-map. The proposed laser profiling system shows the ability to identify and quantify the depth of corrosion pits which can be a very useful parameter for carrying out maintenance. Fig. \ref{fig:rgb_metal_mesh} shows the RGB-D colour map with visual information for qualitative assessment of colour mapping. Fig. \ref{fig:rgb_metal_wire} shows the wire mesh of the RGB colour mapped 3D mesh. 

% \begin{figure}[H]
%     \centering
%     \includegraphics[width=\linewidth]{rgb_metal_wire.png}
%     \caption{3D mesh profile of the coloured pipe surface}
%     \label{fig:rgb_metal_wire}
% \end{figure}

\subsection{Real-time performance}
\begin{figure}[H]
    \centering
    \includegraphics[width=1.3in]{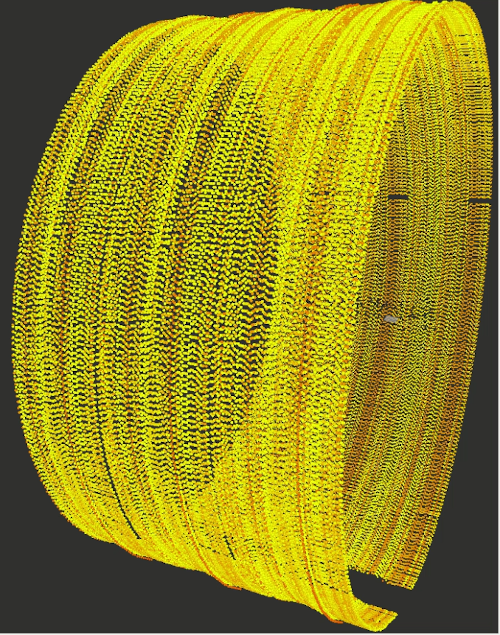}
    \caption{A real-time visualization of the pipe scan in 3D}
    \label{fig:rviz_display}
\end{figure}

Fig. \ref{fig:rviz_display} shows the  real-time 3D point cloud generation in ROS RVIZ for a pipe with liner defects and joints. The hardware performance in combination with the optimized algorithms and parallel processing ROS nodes; the system is capable of processing the image frames without any lag (30 frames/second rate). In the prototype system with a 30 fps camera the robot needs to travel 0.2m per minute to acquire 1mm accuracy along the pipe axis. Higher speeds contributes to lower accuracies however, wouldn't affect the real-time system performance. Table \ref{tab:performance} presents  system's performance for different point cloud resolutions. %Hardware specifications are  given in \ref{hardware}.

\begin{table}[H]
    \centering
    \caption{System performance}
    \begin{tabular}{|c|c|}
        \hline
        \textbf{Points per frame} & \textbf{Frame rate (fps)} \\
        \hline
        1000 & 60 \\
        \hline
        3000 & 30 \\
        \hline
        6000 & 10 \\
        \hline
    \end{tabular}
    \label{tab:performance}
\end{table}

\section{CONCLUSION}
In summary, this paper proposed a 3D laser  profiling in pipelines using stereo vision and structured IR laser light. The performance and the  accuracy of the proposed system have been improved considerably  by employing  image processing algorithms  such as stereo mapping, ray casting, and RGB depth mapping in a  processing pipeline. The system is capable of generating an accurate 3D point cloud of the  internal surface  of the pipe  in real-time. The accuracy of the proposed system is validated on controlled pipe samples. Finally, the tests were carried out by scanning corroded pipe samples to identify defects. Further research is planned to improve the localization of the robot where it is planned to deploy on real pipeline  to test its performance and accuracy. 

\section*{ACKNOWLEDGMENT}
% Acknowledge Sydney Water. Get the exact wording from Karthick

%%%%%%%%%%%%%%%%%%%%%%%%%%%%%%%%%%%%%%%%%%%%%%%%%%%%%%%%%%%%%%%%%%%%%%%%%%%%%%%%

This publication is an outcome from the project ``Development of sensor suites and robotic deployment strategies for condition assessment of concrete sewer walls" funded by the Sydney Water Corporation. 

% \bibliographystyle{apacite}
%\printbibliography

\ifCLASSOPTIONcaptionsoff
  \newpage
\fi

   \bibliographystyle{IEEEtran}
 
\bibliography{IEEEabrv,Reference}
\end{document}